\documentclass[runningheads]{llncs}
\usepackage{graphicx}

\usepackage{tikz}
\usepackage{comment}
\usepackage{amsmath,amssymb} 
\usepackage{color}

\usepackage[accsupp]{axessibility}  

\usepackage{booktabs}
\usepackage{times}
\usepackage{epsfig}
\usepackage{cite}
\usepackage{bbding}
\usepackage{tablefootnote}
\usepackage{overpic}
\usepackage{enumitem} 
\usepackage{overpic} 

\usepackage{xspace}

\newcommand{\todo}[1]{\ClassWarning{NOT READY TO SUBMIT}{There is something left todo}
\textcolor{blue}{[TODO: #1 ]}}
\newcommand{\name}{RadSegNet\xspace}

\begin{document}
\pagestyle{headings}
\mainmatter
\def\ECCVSubNumber{3236}  

\titlerunning{RadSegNet}
%
\author{Kshitiz Bansal$^*$\and
Keshav Rungta$^*$\and
Dinesh Bharadia}
\authorrunning{K. Bansal et al.}
%
\institute{University of California, San Diego\\
\url{ksbansal@eng.ucsd.edu, krungta@ucsd.edu, dineshb@eng.ucsd.edu}}


\title{RadSegNet: A Reliable Approach to Radar Camera Fusion}

\maketitle
\begin{abstract}
   Perception systems for autonomous driving have seen significant advancements in their performance over last few years. However, these systems struggle to show robustness in extreme weather conditions because sensors like lidars and cameras, which are the primary sensors in a sensor suite, see a decline in performance under these conditions. In order to solve this problem, camera-radar fusion systems provide a unique opportunity for all weather reliable high quality perception. Cameras provides rich semantic information while radars can work through occlusions and in all weather conditions. In this work we show that the state-of-the-art fusion methods perform poorly when camera input is degraded, which essentially results in losing the all-weather reliability they set out to achieve. Contrary to these approaches, we propose a new method, RadSegNet, that uses a new design philosophy of independent information extraction and truly achieves reliability in all conditions including occlusions and adverse weather. We develop and validate our proposed system on the benchmark Astyx dataset and further verify these results on the RADIATE dataset. When compared to state-of-the-art methods, RadSegNet achieves a 27\% improvement on Astyx and 41.46\% increase on RADIATE, in average precision score and maintains a significantly better performance in adverse weather conditions.

\end{abstract}


\section{Introduction}


\def\thefootnote{*}\footnotetext{These authors contributed equally to this work}\def\thefootnote{\arabic{footnote}}
Rapid research in self-driving sensing systems has significantly improved the quality of perception tasks like object detection in the past few years. Despite these advancements, we still do not see a prevalence of level 4 or 5 self-driving ability in commercial vehicles. The primary reason behind autonomous cars not being more commonplace is their dependence on lidars, cameras, or their fusion, which are unable to perform robustly in cases of occlusions and adverse weather conditions~\cite{waymo_last,feng2020deep}. This shortcoming, in cameras and lidar, has sparked a major interest in automotive radar-based sensing, particularly in camera radar fusion systems~\cite{nabati2020centerfusion, kim2020low,nobis2019deep}. 

A camera radar fusion system ideally can combine the benefits of both cameras and radars while also addressing the shortcomings of each sensor. While cameras provide rich texture and semantic information, they start failing in case of long range, occluded objects and adverse weather conditions \cite{feng2020deep, chadwick2019distant}. On the other hand, radars are capable of providing all-weather reliance, long range and occlusion-free detection \cite{feng2020deep,palffy2019occlusion,holder2018measurements}; but, they struggle in clearly identifying objects due to lack of rich texture and semantic features \cite{bansal2020pointillism}. In this work, the primary question we try to answer is, how do we fully realise the advantages of both modalities for achieving accurate and also reliable object detection.

\begin{figure}[t!]
    \centering
    \includegraphics[width=0.8\textwidth]{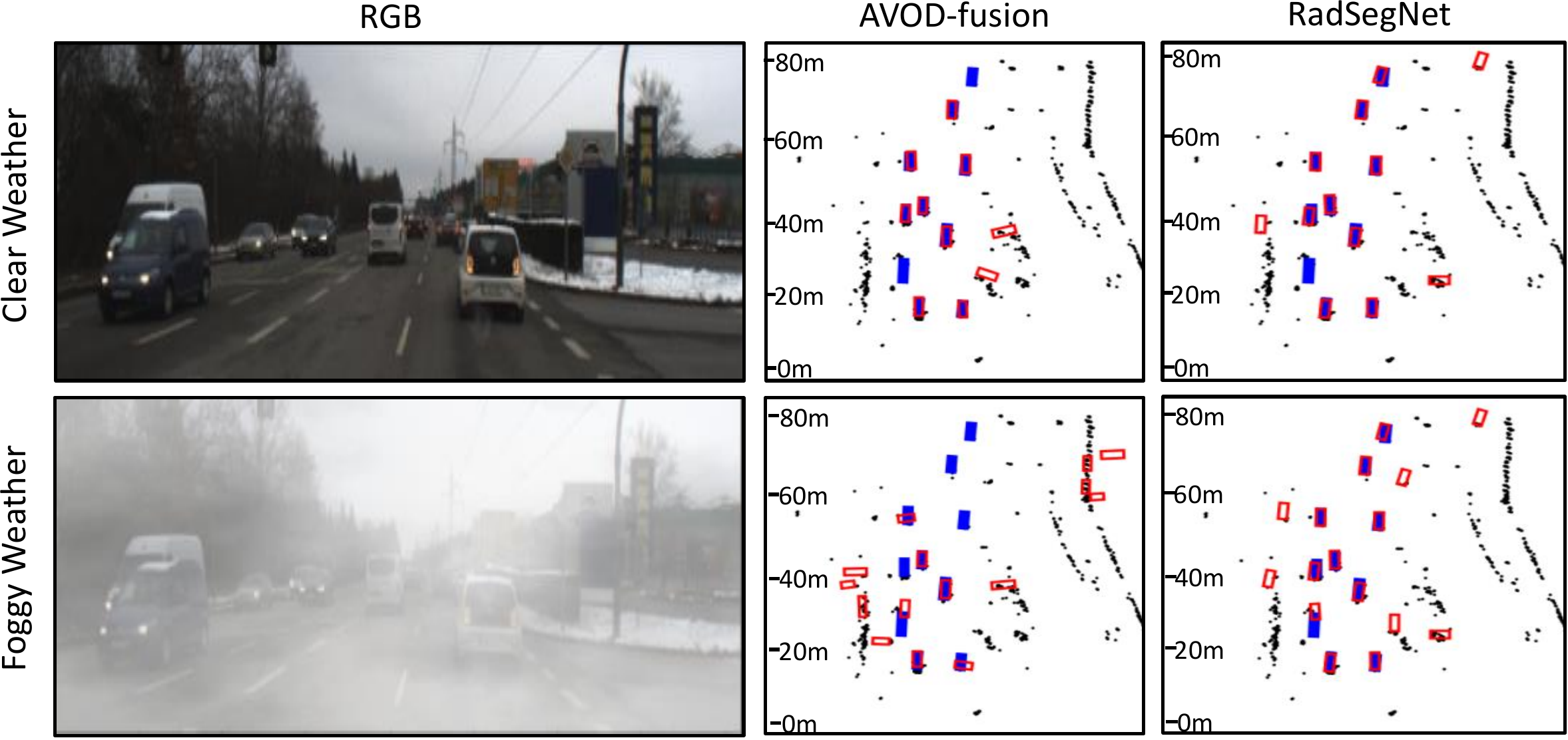}
    \caption{Performance of radar-camera fusion architectures when camera input is augmented with artificial fog. AVOD-fusion~\cite{kim2020low} gets significantly deteriorated while our method continues to provide robust results even in fog. Filled blue boxes \textcolor{blue}{$\blacksquare$} are ground truth and red empty boxes \textcolor{red}{$\square$} are prediction results.}
    \label{fig:bev_rgb}
\end{figure}

An ideal fusion system has to realise the benefits of both sensing modalities, and at the same time, also ensure that the shortcomings of one sensor does not affect the performance of the other. Past works in radar-camera fusion have used projection of radar data on camera perspective view, but operating in this view limits the performance in cases like occlusion of objects \cite{wang2019pseudo}. This results into radars not being used to their fullest. More advanced and state-of-the-art approaches perform fusion at feature level. For example AVOD-fusion~\cite{kim2020low} first simultaneously extracts features from camera perspective and radar bird's eye view (BEV) and then fuses them on per object basis, to take advantage of both the sensors in their views. However, we find that the simultaneous feature extraction and fusion approach does not account for the cases where camera become unreliable. For example, in cases of occluded objects or adverse conditions, radars remain unaffected, but camera input can be highly unreliable. This causes a significant performance loss of the entire system. The loss in performance is illustrated in figure \ref{fig:bev_rgb}, (3rd column), where the quality of detections from AVOD-fusion degrade, when the camera is subjected to artificially generated adverse weather.

Clearly, there is need to improve the reliability of camera-radar systems to achieve good performance even when camera input is degraded. In order to achieve this goal, we argue for a fundamentally different approach for fusing information from radars and cameras. We propose that if we can extract the useful information from both the sensors independently, then we would get the advantages of both modalities without degrading either in case of unreliability. This new philosophy of fusion uses the fact that camera and radar provide complimentary qualities i.e., rich texture and semantic information from cameras can be used to identify objects in the scene, while long range, occlusion free and adverse weather reliable detections can be achieved by using radars. Hence, extracting the information independently is possible which would benefit the reliability of the system.  

The natural question is, how do we design a system that independently extracts the useful information from both modalities. In good conditions the system should be able to use the rich texture and semantic information from camera as well as the useful information like depth and size of all objects from radars, while in case of unreliability in camera caused by occluded or distant object in good weather or image degradation in adverse weather, system should still be able to use radar data reliably. For realising such a system, we propose \name that achieves the required functionality by using mainly two design principles. First principle is based on the insight that for radars, BEV representation offers several advantages over perspective view~\cite{wang2019pseudo} especially in case of occlusion. Hence, at its core, \name uses the radar's BEV representation for detection, to encode all the information present in radar. Next, we note that rich textural and semantic information in cameras is mainly used to clearly identify the objects of a scene. Hence, inspired from~\cite{vora2020pointpainting}, we bank on the significant advancements made in the camera semantic segmentation literature and independently extract the semantic features from camera RGB images. 

However, propagating the semantic information extracted from camera, to radar data, is still a challenging task, as the camera does not have depth information. To overcome this challenge, \name creates a novel semantic-point-grid (SPG) representation to encode the semantic information from camera images into the radar point clouds. For associating the semantics to the radar points, SPG finds the camera pixel correspondences for each radar point, instead of projecting camera image to the radar BEV. Thus, SPG encoding achieves the required independent extraction of information, by distilling the information from camera, adding it to the radar and performing detection on this augmented radar representation. Even in the conditions when camera input is unreliable, \name continues to work reliably using the radar data. Note that these conditions include adverse weather as well as occlusions and long range in clear weather, where camera data can become unreliable. 

We evaluate our approach on two publicly available radar datasets with different types of radars. For comprehensive benchmarking we use Astyx~\cite{meyer2019automotive} dataset that has radar point clouds and augment it for bad weather. For adverse weather testing on real world data, we use the RADIATE~\cite{sheeny2020radiate} dataset that has dense radar data. For the task of object detection, \name sees an average precision (AP) improvement of 27\% in Astyx dataset and 41.16\% in RADIATE dataset, when compared to the state-of-the-art (SOTA) approach of camera-radar fusion, AVOD-fusion~\cite{kim2020low}. More importantly, we show that even in cases where the input from the camera is unreliable, our approach provides much more reliable performance than SOTA. In cases of adverse weather conditions, where camera images are poorer, SOTA could see more than 50\% AP degradation, while, \name degrades less than 6\%. Figure \ref{fig:bev_rgb} shows the robust performance of our approach compared to AVOD-fusion~\cite{kim2020low} when fog is inserted into the images from Astyx dataset. To summarize, our approach to radar-camera fusion is universal (works with any type of radar), reliable (maintains reliability in all-weather conditions), and complete (completely use all advantages of radar sensing). It can also be readily integrated with other sensor-fusion approaches as it does not have any dependency in feature extraction stage. 

\section{Related Work} \label{sec:related}
Current radar and camera fusion approaches can be classified into one of the following categories: projection based (perspective projection or inverse projection or radar-based region proposals), multi-view feature aggregation, and uncertainty based fusion.

    \subsubsection{Radar to camera projection} A common way to do camera-radar fusion is to project radar 3D points to camera 2D perspective view using camera matrices. Felix et al. \cite{nobis2019deep, nabati2020centerfusion} project the radar point on the camera plane and augment them into vertical lines and pillars respectively to encode height. Chang et al.~\cite{chang2020spatial} use a spatial attention network to process projected radar image. Grimm et al.~\cite{grimm2020warping} use a differentiable warping function to warp radar tensor to camera image for using camera labels for training. However, operating in perspective view makes it harder to distinguish between small objects close to the sensor and large objects at a longer range, hence achieving sub-optimal performance.
    
    \subsubsection{Camera to Radar inverse mapping} Another way to fuse data is by projecting camera to radar's bird's eye view. Lim et al.~\cite{lim2019radar} use planar homography transformation to project camera image to radar BEV. However, an inverse mapping of camera to BEV plane is ill-defined due to the lack of depth information in camera images causing ambiguities in detection.
    
    \subsubsection{Region proposal} This class of methods tend to use radars for generating region proposals for performing object detection. Nabati et al.~\cite{nabati2019rrpn} use radars to generate 2D proposals for object detectors like Faster-RCNN that improve the performance for autonomous driving cases. The authors further extends the approach to 3D proposals and utilizes both radar and camera features to refine the proposals and detection~\cite{nabati2020radar}. These methods also operate in perspective view which makes the detection task harder.
    
    \subsubsection{Sensor uncertainty based fusion} Kowol et al.~\cite{kowol2020yodar} use radar detections to generate an uncertainty measure. This measure is used to prune the 2D predictions generated by standard object detectors like Faster RCNN to improve their performance. However, they only use radar to assist camera, thereby not utilising its complete advantages.
    
    \subsubsection{Multi-view feature aggregation} Kim et al.~\cite{kim2020low} use an AVOD \cite{ku2018joint} type architecture where a set of proposals are projected independently to camera and radar plane to extract features. Object proposal wise features are fused together to obtain final predictions. These approaches yield SOTA results for camera-radar fusion but the performance is not reliable in adverse conditions.  

\name uses the SPG representation, which combines the benefits of BEV with camera information in a reliable way. This allows \name to achieve more accurate results along with reliablity in all conditions.


\section{Radar Primer} \label{sec:background} 

At the core, radars use the same time of flight (ToF) analysis of reflections to generate points, as in LiDARs, but they differ in the wavelength of operation. While, lidars use nanometer wavelength signals, which provide them with very high resolution due to surface scattering, radars use millimeter wavelengths where the reflected power is divided between specular reflections and diffused scattering~\cite{bansal2020pointillism}. The raw radar data is dense, but contains background thermal or multipath noise~\cite{barnes2020oxford}. Radar data is also commonly subjected to Constant False Alarm Rate (CFAR) filtering, which generates a light-weight and sparse point cloud output~\cite{meyer2019automotive}. Consequently, object edges are not as clearly defined in radar point clouds as they are in LiDAR point clouds. For example, in radar point clouds, a point cluster originating from a wall could have a similar spatial spread as that of a cluster originating from a car. This effect makes it challenging to learn any shape based features directly from radar point clouds to distinguish objects of interest (Cars, pedestrians, etc) from background objects. Figure \ref{fig:bev_rgb} shows the non-uniformity present in radar point clouds.  
 
However, at the same time, radars also offer the following unique advantages due to the millimeter band transmission: a) They provide a \textit{longer range} than lidars, because higher wavelength signal has a lower free space power decay rate. This allows radar waves to travel over longer distances \cite{radar-vs-lidar,feng2020deep}. b) They can \textit{see-through occluding vehicles} because their signals bounce off the ground allowing them to also sense vehicles which are completely occluded~\cite{palffy2019occlusion,holder2018measurements}. c) they are an \textit{all weather sensor} because the larger wavelength of millimeter waves allows them to pass unaffected through adverse conditions like fog, snow and rain.

\begin{figure*}[t!]
    \centering
    \includegraphics[width=\linewidth]{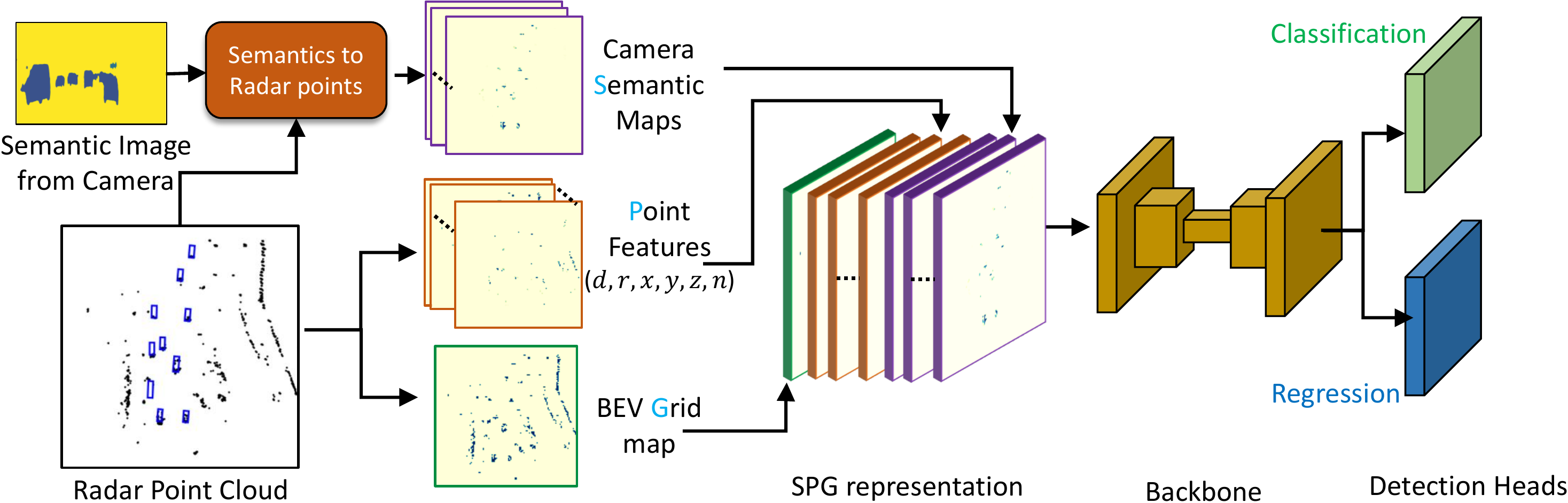}
    \caption{Overview RadSegNet: Our approach utilizes encodings from our SPG module to detect objects. The encodings are generated from the semantic features from a semantic segmentation network along with radar point-based features and an occupancy grid. These encoded maps are concatenated and passed through the bounding box detection network.}
    \label{fig:overview}
\end{figure*}

\section{Methodology}
In this section we break down \name's architecture into its various stages, explaining how \name is able to tackle challenges like occlusion and all weather reliance by utilising the independent feature extraction philosophy of fusion.



\subsection{BEV input representation} \label{lab:BEV_vs_Persp}
 The view used to represent input data has a significant impact on deep learning architecture's performance for object detection tasks. Wang et al.\cite{wang2019pseudo} show that performance gains can be obtained by just transforming data from a perspective camera view to a 3D/BEV view. The reason behind this is that in perspective view, there is scale ambiguity with depth as well as object overlap due to occlusions. Local computations like 2D convolutions, on a 2D perspective view image, can cause objects at different depths to be processed with the same kernel. This makes the task of object detection much harder to learn. BEV representation, on the other hand, is able to clearly separate objects at different depths, offering a clear advantage in cases of partially and completely occluded objects~\cite{wang2019pseudo}.
 
Our key insight is that for radars, BEV becomes an absolute necessity, as they even get signals from occluded objects due to the radio waves bouncing off of the ground (section~\ref{sec:background}). Representing radars in perspective view to extract features is not only sub-optimal, but may also cause confusion in case of occluded objects. Hence, for getting a good and reliable performance, \name uses BEV representation as an input.

\subsubsection{BEV Occupancy grid}
In order to generate a BEV representation, we project the radar points onto a 2D plane by collapsing the height dimension. The plane is then discretized into an occupancy grid. Each grid element is an indicator variable that gets a value of 1 if it contains a radar point otherwise it is represented as 0. This BEV occupancy grid also preserves the spatial relationships between the different points of an unordered point cloud and stores radar data in a more structured format \cite{lang2019pointpillars}. 

\subsubsection{Radar point features}
The BEV occupancy grid provides an optimal representation for radars and provides order to the un-ordered radar point cloud. However, a BEV grid also discretizes the sensing space into grids which dissolves the useful information required for the refinement of bounding boxes. To retain that information we add the point based features to our BEV grid as additional channels. Specifically, we add the cartesian coordinates, doppler and intensity information as additional features. The BEV grid input to the network is then defined as follows:
\begin{gather}
    s := (\mathcal{I}((u,v)), d, r, x, z, y, n)
\end{gather}

Here, $\mathcal{I}$ represents the 2D occupancy grid where each grid element is parameterized as $(u,v)$. All the positions in $\mathcal{I}((u,v))$ where radar points are present store 1 or else 0. $d$ and $r$ represents the doppler and intensity value of radar points. They help identify objects based on their speeds and reflection characteristics. $(x, z)$ is the average depth and horizontal coordinate in the radar's coordinate system. In order to encode height information, we generate height histograms by binning the height dimension ($y$) at 7 different height levels and creating 7 channels, one for each height bin. The cartesian coordinates $(x,y,z)$ help in refining the predicted bounding box. The $n$ channel contains the number of points present in that grid element. The value of $n$ is usually proportional to the surface area and reflected power which helps in refining bounding boxes. An overview of all the point features is also shown in figure~\ref{fig:overview}.


\subsection{Fusion with Camera Data}
The BEV occupancy grid, along with the radar point features, represents all the information in radar point clouds in a well structured format. Now, we need to add camera information to this representation to complete our fusion system. Note that a direct projection of camera data to the BEV is non-trivial and challenging as camera lacks depth information. To solve this problem, current state-of-the-art radar-camera fusion systems~\cite{kim2020low} simultaneously extract features from both modalities. The features are then fused on a per-object proposal basis. However, in this approach, the performance drops significantly when the camera data is unreliable for any object, in case of occlusion or adverse weather (drop of more than 50\% in some cases, refer section~\ref{sec:results}). 

In \name, we define a novel SPG (Semantic-point-grid) encoding that solves the above challenge by independently extracting information from cameras, in a reliable way. Our SPG encoding first distills the rich texture and semantic information from cameras and combines it with radar point clouds. In the next section, we provide details of our SPG encoding and how it makes use of all the advantages present in both modalities while being reliable in cases of camera uncertainty. 

\subsection{Semantic-Point-Grid feature encoding}\label{sec:bev_map_gen}


\subsubsection{Camera semantic features}

The rich texture and semantic information in camera images is very useful for understanding a scene and identifying the objects in it. This information complements well with the radar, where non-uniformity in point clouds makes it harder to learn features that can identify the objects well (section~\ref{sec:background}). Our key insight for using this complementary nature while maintaining reliability in adverse conditions, is to first extract the useful information from camera images in the form of scene semantics and then use it to augment the BEV representation obtained from radar. In contrast to fusing the features on a per-object basis, our approach keeps a clear separation between information extraction from two modalities, hence performing reliably even when one input is degraded. We use a robust pre-trained semantic segmentation network to obtain semantic masks from camera images of the objects present in the scene. However, we still need to add this information to the radar BEV without the presence of depth information for camera image. 

\subsubsection{Adding semantics to SPG}
\begin{figure}[t!]
    \centering
    \includegraphics[width=0.5\linewidth]{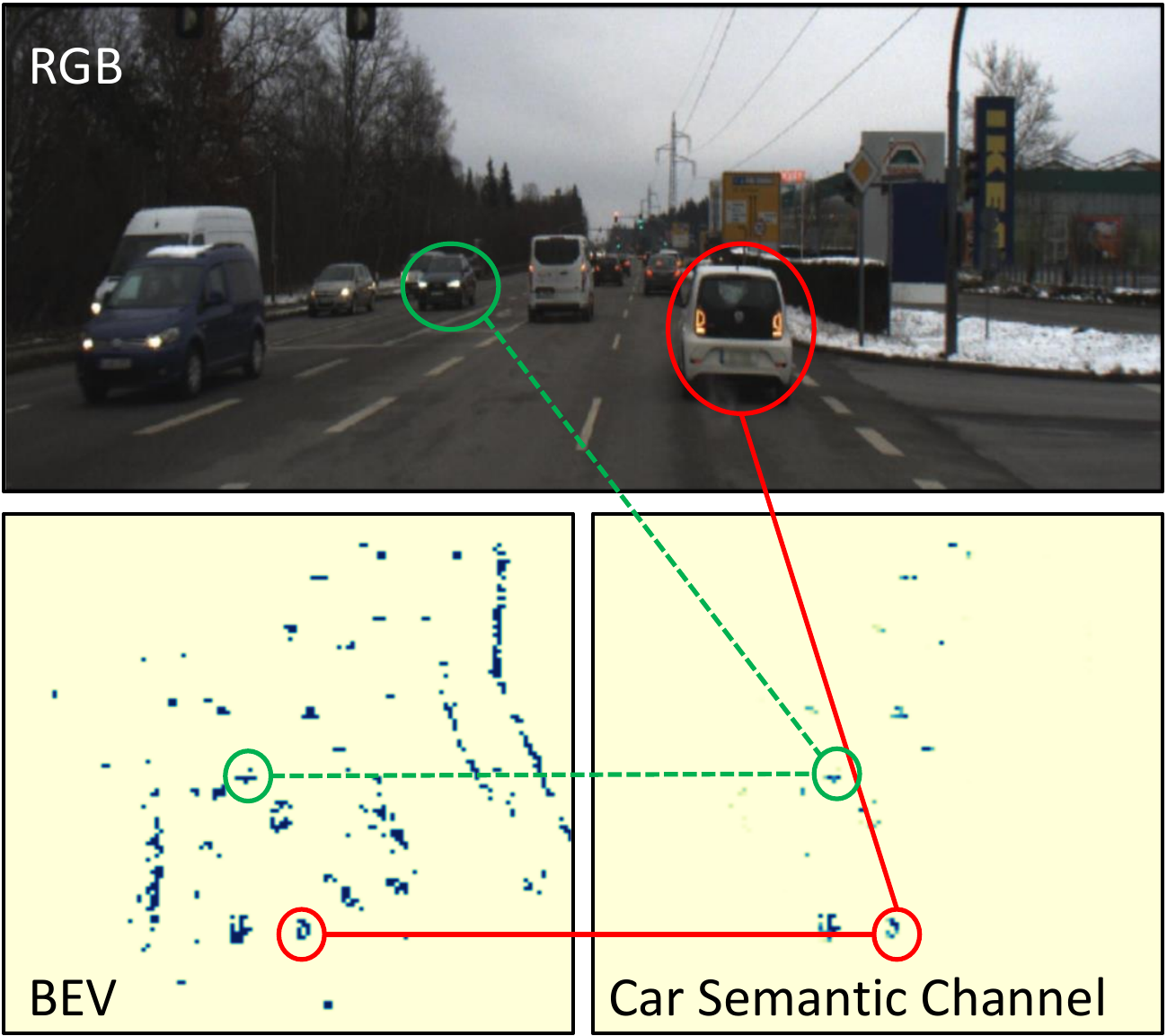}
    \caption{Adding the semantic channel in SPG encoding helps in identifying points belonging to objects of interest. This figure shows how two cars are represented in the semantic map for the class cars and in the corresponding BEV occupancy grid. Similarly semantic maps can be represented for other classes.}
    \label{fig:semantic_map}
\end{figure}

To associate camera based semantics to radar points, we create separate maps for each output object class of the semantic segmentation network. These maps are of the same size as the BEV occupancy grid and get appended as semantic feature channels. To obtain the values of the semantic feature channels for each grid element, we first transform the radar points to the camera coordinates. Next, we find the nearest pixel in camera image to the transformed point and use the semantic segmentation output of that pixel as the values of semantic feature channels in SPG. In case multiple radar points belong to same grid element, an average is taken over all the resulting semantic values. These feature channels contain the semantic information extracted from the camera, helping in the object detection from radar BEV occupancy grid. They effectively reduce the possible false positive predictions generated by radars, as the radars may get confused in identifying objects due to inherent non-uniformity (section \ref{sec:background}) in radar data. Figure \ref{fig:semantic_map} shows and example of how the semantic features are encoded with the radar BEV grid, for the class car. Figure \ref{fig:overview} shows the overview of entire \name.

Note that the form of fusion with camera, used in \name, does not filter out any radar points, while making better use of the advantages that both modalities bring to the table. This means that in cases where the camera based features become less informative, all objects in the scene are still visible to radars, which prevents any drastic drop in performance. The textural and high resolution information from cameras is condensed into semantic features which assists the all-weather, long range and occlusion robust sensing of radars. 

\subsection{Bounding Box prediction on SPG features}\label{sec:feature_extraction_&_detection}

Each of the BEV maps generated through our SPG encoding, are passed into a deep neural network for feature extraction and bounding box prediction. For our backbone feature extraction, we use an encoder-decoder network with skip connections. We use 4 stages of down-sampling layers with 3 convolutional layers at each stage to extract features of different scales during the encoding stage and then combine all the intermediate features during the up-sampling stage by using skip connections to generate the final set of features. We use an anchor box-based detection \cite{liu2016ssd} architecture to generate predictions using a classification and a regression head. The classification head predicts a confidence score for the output boxes and a regression head learns to refine their dimensions.





\section{Implementation}

\subsubsection{Image segmentation network} For our image segmentation network, we use a pre-trained semantic segmentation model from the model zoo provided by the official DeeplabV3+ implementation~\cite{chen2017rethinking,chen2018encoder}. We use the ResNet-101 model~\cite{he2016deep} which is trained on the Cityscapes dataset \cite{cordts2016cityscapes} for semantic segmentation task. We choose this model for its accuracy and generalizability. However, depending upon usage an alternative model optimized for speed can also be chosen. Our approach is agnostic to the type of network we choose.

\subsubsection{Loss functions}
In this architecture, we are using a combination of two-loss functions as our objective to train the network. The classification head uses a focal loss \cite{lin2017focal} which provides better results for sparse radar point clouds than binary cross entropy. For regression head we use a Smooth L1 loss which combines L1 and L2 losses. The losses are given by:
\vspace{-5pt}
\begin{gather}
    L_{foc} = -\alpha(1-p_t)^{\gamma}\log(p_t)\\
    L_{Smooth_{L1}} = \begin{cases}
    0.5\sigma^2\Delta^2 & |\Delta| < \frac{1}{\sigma^2}\\
    |\Delta| - \frac{0.5}{\sigma^2} & else
    \end{cases}\\
    L_{tot} = L_{foc} + L_{Smooth_{L1}}
\end{gather}

where $p_t$ is the confidence output of classification head, $\Delta$ is the refinement value and $\alpha, \gamma, \sigma^2$ are hyper-parameters of loss functions. 

\subsubsection{Training Details} For each frame in the datasets, the radar data is processed to extract the initial feature channels. The input into the network is a $(N,C,W,H)$ tensor where $N=2, C=22, W=128, H=128$. The channels correspond to semantic segmentation values (9), BEV occupancy grid map (1) and point features (12). We use BEV ground-truth labels to train the classification and regression head. We use the average dimension of the ground truth labels as our fixed anchor box sizes. We use a target IoU (Intersection over Union) of 0.5 to determine the positive and negative examples of the anchor boxes for classification. Only the boxes marked as positive examples are used for regression loss.

The values of our hyper-parameters are ascertained empirically. These are: $\alpha = 0.9, \gamma = 2.0, target_{IoU} = 0.5, \frac{1}{\sigma^2} = 1$. Our network is trained using a learning rate of $\lambda = 0.001$ and a weight decay of $\eta = 1e-5$ for the Adam Optimiser. We train our network for around 20 hours using 2 GTX 1080TIs and batch size of 2, to reach convergence and use early stopping to evaluate the system using the best model. We perform k-fold cross validation to ensure better generalizability.

\subsubsection{Metrics}
We use BEV average precision (AP) as our main metric in our evaluation. AP is defined for a particular Intersection over Union (IoU) threshold of a predicted bounding box with the ground truth box. We use IoU threshold of 0.5 in our evaluation to determine True Positives, which is commonly used across all BEV object detection benchmarks \cite{geiger2012we}.

\subsubsection{Perspective view Baseline} We choose CenterFusion\cite{nabati2020centerfusion}, the state-of-the-art perspective view based camera radar fusion approach, as one of our baselines. In this approach, the authors create a feature map of radar point clouds and process it along with the corresponding image based feature map to perform detections. We also compare our approach against a camera only approach called CenterNet\cite{zhou2019objects}. CenterNet is essentially CenterFusion without the corresponding radar data. We use the official GitHub implementation of these networks. We take the pre-trained network provided by the authors and fine-tune it on the Astyx dataset in order to make it a fair comparison. The pre-trained network performed better than training the network from scratch on the Astyx dataset. Hence, we provided only the results for the fine-tuned network.

\subsubsection{Multi-view Baseline (SOTA)} We use \cite{kim2020low} as our multi-view aggregation based baseline. \cite{kim2020low} uses an AVOD~\cite{ku2018joint} architecture to perform radar-camera fusion. Due to the unavailability of official code from \cite{kim2020low} we use the official implementation of AVOD and train it on the Astyx dataset to compare the performance. We dub this approach as AVOD-fusion. This is also the SOTA approach for sensor fusion.

\subsubsection{Testing datasets} We perform an evaluation on two datasets. First we show results on Astyx Hi-res radar dataset~\cite{meyer2019automotive} and comprehensively benchmark our approach. We also create augmented weather in this dataset to evaluate the reliability of different camera-radar fusion approaches in adverse conditions. Next, we evaluate on RADIATE dataset~\cite{sheeny2020radiate} which contains images from real world bad weather environment.

\section{Evaluation on Astyx Dataset}\label{sec:results} 
In this section, we provide a comprehensive evaluation of our system on publicly available Astyx dataset \cite{meyer2019automotive} to compare our system with multiple different baselines.

\subsubsection{Dataset Details}
Astyx Hi-res radar dataset~\cite{meyer2019automotive} is the only publicly available dataset with a high-resolution MIMO radar that provides point clouds. It was collected on the roads of Germany with a vehicle moving at different speeds. There are a total of 546 frames provided in the set. The radar data is in the form of a point cloud where each radar point consists of (x,y,z) location, a doppler estimate and an intensity estimate. The dataset contains 3D bounding box labels of vehicles and pedestrians, generated via human annotation using an onboard lidar point cloud and camera images. For each label, in addition to the position, dimension and orientation of the object, the level of occlusion is also provided. We evaluate the dataset in 3 categories "No occlusion (Easy)", "Not fully occluded (Medium)" and "Full Dataset (Hard)" based on the occlusion level of the objects. We evaluate AP performance for task of vehicle detection (cars and trucks). We split the dataset using a 4:1 ratio for the training and test sets. Most of the labels are present within 80m distance from radar as lidars are unable to maintain enough point density at such large distances, causing label certainties to degrade drastically beyond this limit. Hence, we limit all evaluations of our system to that distance.






\begin{table*}[t!]

\caption{BEV Average Precision scores for different IoU thresholds in brackets. \name outperforms the baseline architectures consistently over all difficulty levels. Scores for the best baseline are underlined. R: Radar; C: Camera}
\centering

\begin{tabular}{|c|c||c|c|c|}
\hline
Method & Modality & \multicolumn{3}{c|}{AP (0.5)} \\ \hline
      &          &    Easy     & Medium     & Hard    \\ \hline
CenterNet~\cite{zhou2019objects}        & C    & 12.40 & 13.36  & 13.11 \\ \hline
Center Fusion~\cite{nabati2020centerfusion}    & R + C   & 9.78 & 11.22 & 10.87 \\ 
Painted-pointpillars~\cite{vora2020pointpainting} & R + C & -- & -- & 36.00  \\
AVOD-Fusion~\cite{kim2020low}             & R + C   & \underline{40.38} & \underline{37.46} & \underline{36.11}\\ \hline

\name (Ours) & R + C   &  \textbf{48.14}  & \textbf{46.82}    & \textbf{45.88}  \\ 
\% increase over \underline{underlined} &    &  \textcolor{blue}{+19.21\%} & \textcolor{blue}{+25.00\%}   & \textcolor{blue}{+27.05\%} \\ \hline 

\end{tabular}%
\label{tab:mAP_performance}
\end{table*}

\subsection{BEV bounding box prediction}

\begin{figure}[t!]
    \centering
    \includegraphics[width=0.8\linewidth]{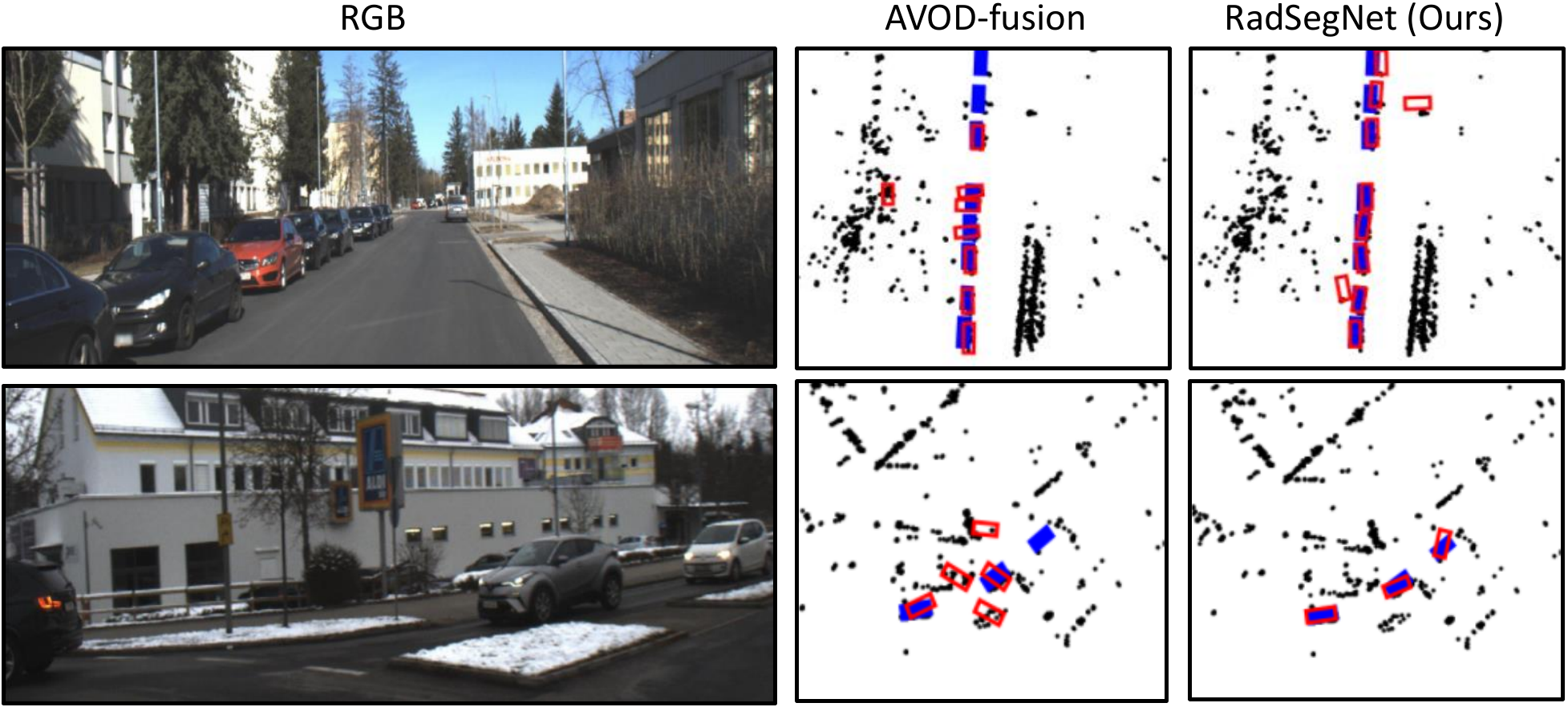}
    \caption{Visualization of the output bounding box output of \name on challenging cases from Astyx Dataset. Black dots represent radar point clouds in BEV. Filled blue boxes \textcolor{blue}{$\blacksquare$} are ground truth and red empty boxes \textcolor{red}{$\square$} are prediction results.}
    \label{fig:qual_astyx}
\end{figure}

Table \ref{tab:mAP_performance} shows the AP score of our network in comparison to other radar-camera fusion approaches. We see that the perspective view based approaches~\cite{zhou2019objects, nabati2020centerfusion} do not provide a good AP score. This shows the superiority of BEV representation which leads to major performance boosts, specially in cases of occlusions and long range. \name outperforms these perspective view baselines in all 3 occlusion categories thanks to its BEV representation. Similarly, the current state-of-the-art approach AVOD-fusion \cite{kim2020low}, which also uses BEV representation from radar, is the best performing baseline. However, \name also outperforms AVOD-fusion across the board in all difficulty categories, showing that independent information extraction provides significant advantages in all conditions. To further analyse this claim, we also provide the percentage increase over AVOD-fusion in all categories. The percentage increase is higher in medium and hard categories that includes occlusion. This shows that the SPG representation for radar-camera fusion in \name, provides significant advantage over simultaneous feature extraction of AVOD-fusion, specially in cases of occlusion, where the camera features are unreliable even in clear weather. We also provide the qualitative output of our network in some sample scenes. Figure~\ref{fig:qual_astyx} shows the bounding box prediction outputs of \name compared to AVOD-fusion. It shows that our network can predict accurate boxes in diverse conditions such as long range, closely spaced cars and different orientations.

\begin{table}[h!]
\caption{Performance comparison between using Lidar and Radar as input}
\centering
\begin{tabular}{|c|c|c|}
\hline
    Model     &  Modality  &  AP (0.5) \\ \hline
Pointpillars~\cite{lang2019pointpillars}  &  L            & 26.74    \\ 
\name-BEV &  L            & 27.11    \\
\name&  L+C          & 36.79    \\ \hline

Pointpillars~\cite{lang2019pointpillars} & R & 31.41  \\ 
\name-BEV     &  R          & 36.09    \\ 
\name     &  R+C           & 45.88    \\ \hline
 
\end{tabular}%

\label{tab:lidar}
\end{table}

\subsection{Performance on Lidar compared to Radar}
In this experiment, we use the liDAR data provided by Astyx as an input to \name without changing the architecture, to understand the advantage of using radars over lidars. For comparison, we also use one of the state-of-the-art lidar object detection networks, pointpillars~\cite{lang2019pointpillars}, with the lidar data provided in Astyx as input. Table~\ref{tab:lidar} provides the comparison results for this experiment. We consider two variants of \name: the complete \name and \name-BEV, where we do not use the semantic features from the camera. For both variants, we compare the performance between using radar and lidar as input. We see that although the lidar based object detection benefits by adding camera (\name vs \name-BEV), it still underperforms compared to using radar as input. Radars provide long range and occlusion free sensing, which significantly benefits the object detection task. Moreover, \name-BEV outperforms pointpillars~\cite{lang2019pointpillars} when compared on radar data, thanks to SPG encoding used in \name, that can encode useful context from radar point clouds. By combining the cameras with radars, \name provides a low cost, all-weather reliable and high quality perception solution. Please refer to supplementary for more qualitative and distance-wise comparison.

 


\begin{table}[h!]
\caption{Effect on AP for Iou threshold 0.5 of different weather conditions on our architecture vs AVOD-fusion. The same input is provided to both the architectures in all the respective experiments. Percentage drop from clear weather performance in parenthesis}
\centering
\begin{tabular}{|c|c|c|c|c|}
\hline
              & \multicolumn{4}{c|}{AP (0.5)}\\ \hline
Model          & Clear & Fog   & Snow  & Rain  \\ \hline
AVOD-Fusion~\cite{kim2020low}      & 36.11 & 2.38 & 33.56 & 17.41\\ 
          & --  & \textcolor{red}{(-93.41$\%$)} & \textcolor{red}{(-7.06$\%$)} & \textcolor{red}{(-51.79$\%$)} \\ \hline
\name & 45.88 & 43.24 & 43.72 & 32.89 \\ 
          & -- & \textcolor{blue}{(-5.75$\%$)} & \textcolor{blue}{(-4.71$\%$)} & \textcolor{blue}{(-28.31$\%$)} \\ \hline
\end{tabular}%

\label{tab:bad_weather}
\end{table}

\subsection{Performance in adversarial scenarios for camera}
In this experiment, we further evaluate the performance of the camera-radar fusion systems when camera images are subjected to adversarial conditions. To compare the performance drop against the normal conditions, we need to augment the camera images in the Astyx dataset with artificial bad weather conditions. Due to the unavailability of dense depth maps and stereo cameras in the Astyx dataset, it is not possible to use physical models of augmentation~\cite{halder2019physics, sakaridis2018semantic}. However, to showcase the proof of concept we use imgaug \footnote{https://imgaug.readthedocs.io/en/latest/source/overview/weather.html} library that uses image filters to add bad weather in the images. Please refer to supplementary material for more details. We also compare results in real adverse weather data in the next section on the RADIATE dataset.

Table \ref{tab:bad_weather} shows the performance comparison of our work against the AVOD-fusion baseline. For each augmented weather condition, we also show the performance drop, as a percentage, from the clear weather performance. We use the network trained on clear weather and evaluate the test set with the augmented weather augmentations. As shown by the results, AVOD-fusion's performance heavily degrades in cases of fog and rain. This is because AVOD-fusion learns feature representations that are heavily dependent on cameras for each object proposal, which becomes reliable in adverse conditions. \name shows that its performance is much less affected in all conditions compared to the AVOD-fusion. These results show the shortcomings of the current radar-camera fusion approaches and the ability of \name to learn independent features from radar and cameras, that can perform reliably in adverse scenarios. Please refer to supplementary material for qualitative comparisons.

We also compare the IoU drop of semantic segmentation output for different augmentations, treating the segmentation output of the original image as ground truth (qualitative outputs in supplementary material). We get an IoU of 0.61 (Fog), 0.40 (Rain) and 0.57 (Snow). The quality degradation follows the same trend as the AP performance. However, past work has shown that the performance of semantic segmentation output can be independently improved by fine tuning on adverse weather data~\cite{sakaridis2021acdc, sakaridis2018semantic}, which would further boost the performance of \name.


 

\vspace{-0.5cm}
\begin{table}[h!]
\centering
\caption{Ablation study experiment. We present the effect of each channel on the overall BEV AP performance of \name at different IoU thresholds}

\begin{tabular}{|c|c|c|c||c|}
\hline
\multicolumn{4}{|c||}{SPG Channel} & \\ \hline
    Radar   &  Position  & Num Pts &   Camera   &  AP (0.5) \\ \hline
 \Checkmark &            &            &             & 32.52 \\ 
 \Checkmark & \Checkmark &            &             & 33.09 \\
 \Checkmark & \Checkmark & \Checkmark &             & 36.09  \\ 
 \Checkmark & \Checkmark & \Checkmark & \Checkmark  & 45.88 \\\hline
 
\end{tabular}%
\vspace{-0.5cm}
\label{tab:ablation}
\end{table}

\subsection{Ablation Studies}
In this section, we evaluate the performance gains provided by each channel of our SPG encoding. Table~\ref{tab:ablation} shows the results of this ablation study. The baseline experiment contains only the BEV map with doppler and intensity features (Radar column). The $(x,y,z)$ (pos) maps provides an improvement of 1.76\% in 0.5 IoU AP score. These channels provide the spatial context of each BEV grid element in the world coordinate system which is specifically helpful in bounding box refinement. The $n$ (Num Pts) channel provides another 9.05\% increase by providing information about the strength and surface area of reflection. Finally, the semantic features from the camera provide a significant 27.12\% increase in performance, which illustrates our claim that the independent information extraction approach used by \name can comprehensively make use of the advantages of both modalities.

\section{Evaluation RADIATE dataset}
\label{sec:radiate_eval}

In this experiment, we evaluate \name on a large scale radar dataset RADIATE~\cite{sheeny2020radiate}. This dataset uses a mechanical radar which provides dense radar data as output. The dataset also contains scenes from adverse weather conditions such as rain and snow, and bad lighting conditions such as night, making it ideal to test the performance on real world adverse conditions.

\subsubsection{Implementation details} RADIATE uses a mechanical Navtech CTS 350-X radar and 2 ZED cameras. The radar data is stored as 2D intensity maps, without any height information. We use the left ZED camera in our evaluation. The ZED camera is only facing forward, so we crop out the intensity maps to only keep the forward direction. The labels are filtered accordingly. The maximum distance of evaluation is about 70.66m. \name uses point clouds as input in order to perform SPG encoding. As the radar input is present in form of intensity maps, we use CFAR~\cite{di2005cfar} filtering technique to convert the intensity maps to 2D point clouds. As there is no height information, we use the height of the sensor as the height coordinate for data, in order to get a 3D point cloud. After this step, we have the camera image and radar point cloud, which we use to evaluate \name on the RADIATE dataset. The training set contains 8890 clear samples and 4151 adverse samples. The test set has 4387 clear and 1222 bad samples. This is the official split provided by the authors~\cite{sheeny2020radiate}.

\vspace{-0.5cm}
\begin{table}[h!]
\caption{Results on the RADIATE dataset. The percentage scores in the bracket show improvement over the clear-only training for respective approaches.}
\centering
\begin{tabular}{|c|c|c|c|c|}
\hline
           &Trained on & Clear          &  Adverse            & Clear + Adverse     \\ \hline
AVOD-fusion& Clear & 43.50          &  18.03          & 36.68           \\ \hline
AVOD-fusion& Clear + Adverse& 44.98          &  22.60 \textcolor{red}{(+25.34\%)}         & 38.17           \\ \hline \hline
\name  & Clear & 58.43 &  17.12 & 47.71  \\ \hline

\name  & Clear + Adverse & \textbf{59.63} &  \textbf{38.70 \textcolor{blue}{(+126.05\%)}} & \textbf{53.92}  \\ \hline
\end{tabular}%
\label{tab:rad_res}
\vspace{-0.5cm}
\end{table}

\subsection{Performance in Clear/Bad Weather}
Table \ref{tab:rad_res} shows the results for the performance of object detection on this dataset. We perform two types of experiments 1) training only on clear samples and 2) training on both clear and adverse samples. The same test set is used for both the experiments, which contains frames from both clear and adverse data. \name achieves better performance compared to the AVOD-fusion baseline in both clear and adverse scenarios (41.46\% improvement when trained on clear+adverse and tested on clear+adverse). More interestingly, we compare the AP score increase seen on adverse weather test set, from first experiment to the second. This increase for \name is much more significant compared to the same for AVOD-fusion (126\% vs 25\%). We make two observations from this: 1) for the dense radar type, there is a slight domain gap between clear and adverse radar data, as a network trained on clear-only does not generalize well over adverse data and 2) \name's approach of independent information extraction provides much more reliable performance than SOTA when some supervision is provided for adverse data. Overall, this experiment further proves that \name's fusion provides better performance in good weather conditions and much more reliability in adverse weather conditions, regardless of the type of the radar.

\begin{figure*}[t!]
    \centering
    \includegraphics[width=0.9\linewidth]{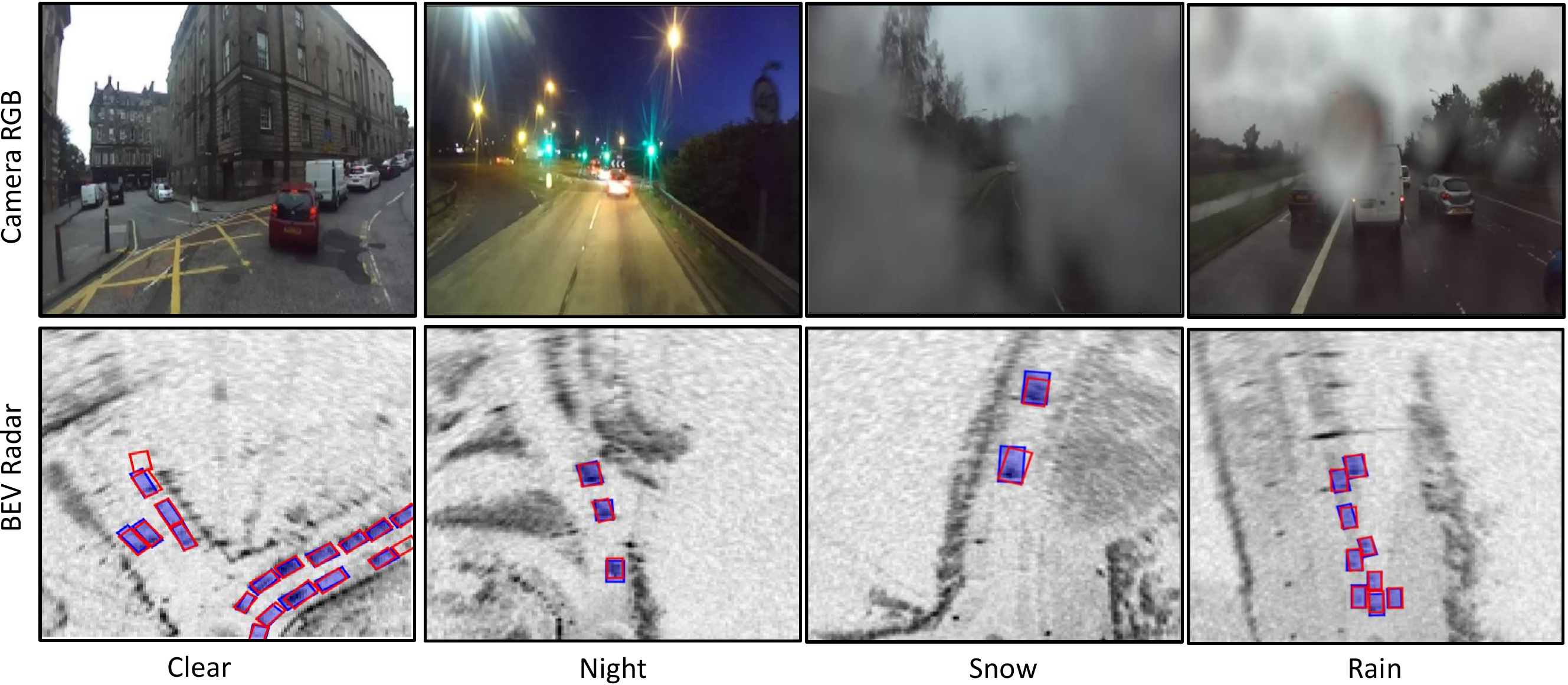}
    \caption{Bounding box prediction results of \name in different weather conditions on RADIATE dataset.~\textcolor{blue}{Blue} box shows the ground truth vehicle and \textcolor{red}{Red} box shows our prediciton.}
    \label{fig:qualitative}
\end{figure*}

Figure \ref{fig:qualitative} provides sample outputs of our network in different weather conditions. The results show that \name's design is agnostic to radar type and generalizes quite well. It provides accurate detections in challenging scenarios of closely spaced vehicles and adverse weather and lighting conditions.

\section{Discussion}
\name performs semantic segmentation camera images and performs detections on SPG encoded representation. To minimize the overhead of obtaining semantic segmentation in a practical scenario, the two steps can run in parallel by keeping one frame latency between the detection and the semantic segmentation network. Past works have explored the possibility of building such systems \cite{vora2020pointpainting} and similar techniques can also be employed in our approach. We showed that independent extraction reduces the co-dependency of camera and radar feature extraction. In the worst cases when camera semantic segmentations are completely degraded, the performance of the entire system would drop down to radar only detections. Future works would be to devise an uncertainty metric that can switch off the camera input after reaching a certain point of degradation. Note that turning off camera is only possible by having independence in camera and radar feature extraction which is provided by \name. Also, the effect of rain, snow, hail and fog on radar has been studied in past literature\cite{zang2019impact}. The overall effect is a rise of noise power at the radar receiver. This effect is fundamentally different from that in lidar and cameras, where bad weather can create false objects and distort the images respectively. An increase in noise level leads to a decrease in maximum range of the radar, which can be undone by using a higher transmit power \cite{zang2019impact} or by doing some fine-tuning on bad-weather data (section~\ref{sec:radiate_eval}).

\bibliographystyle{splncs04}
\bibliography{egbib}

\newpage
\appendix
\chapter*{Supplementary Material}

\section{Overview}
In our paper RadSegNet, we presented a new approach to perform radar and camera fusion that achieves a reliable performance in case of adversities. Our analysis shows, that compared to the state-of-the-art~\cite{kim2020low} approach, performing an independent feature extraction for radar and camera, provides better performance in clear conditions and significantly more reliable performance in adverse weather conditions. To further aid the understanding, we provide additional details and evaluations of our approach in this supplementary document. We first provide details about comparison between lidar and radar input to understand the benefits of using radars. Then, we provide more details about our bad weather experiments (section 6.3) and qualitative visualizations of bounding box prediction on Astyx~\cite{meyer2019automotive} dataset in both good and bad weather to see the effect of weather conditions on object detection. Finally we provide the details of our RadSegNet implementation to ease reproduction.

\section{Performance on Lidar compared to Radar}
\begin{figure*}[t!]
    \centering
    \includegraphics[width=0.5\linewidth]{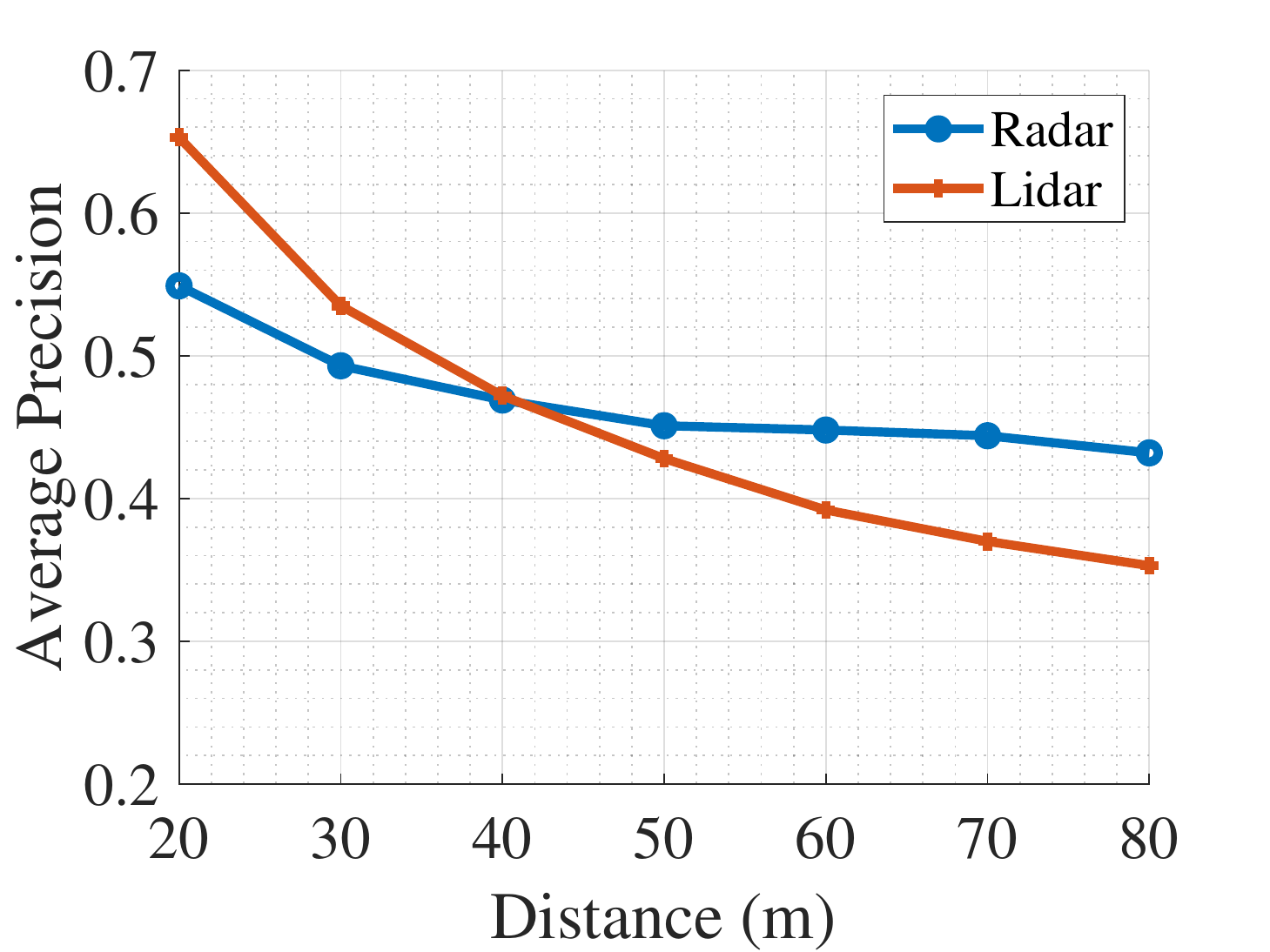}
    \caption{Comparison of bounding box detection performance over distance, between lidar and radar input}
    \label{fig:lidarvsradar}
\end{figure*}

\begin{figure*}[t!]
    \centering
    \includegraphics[width=\linewidth]{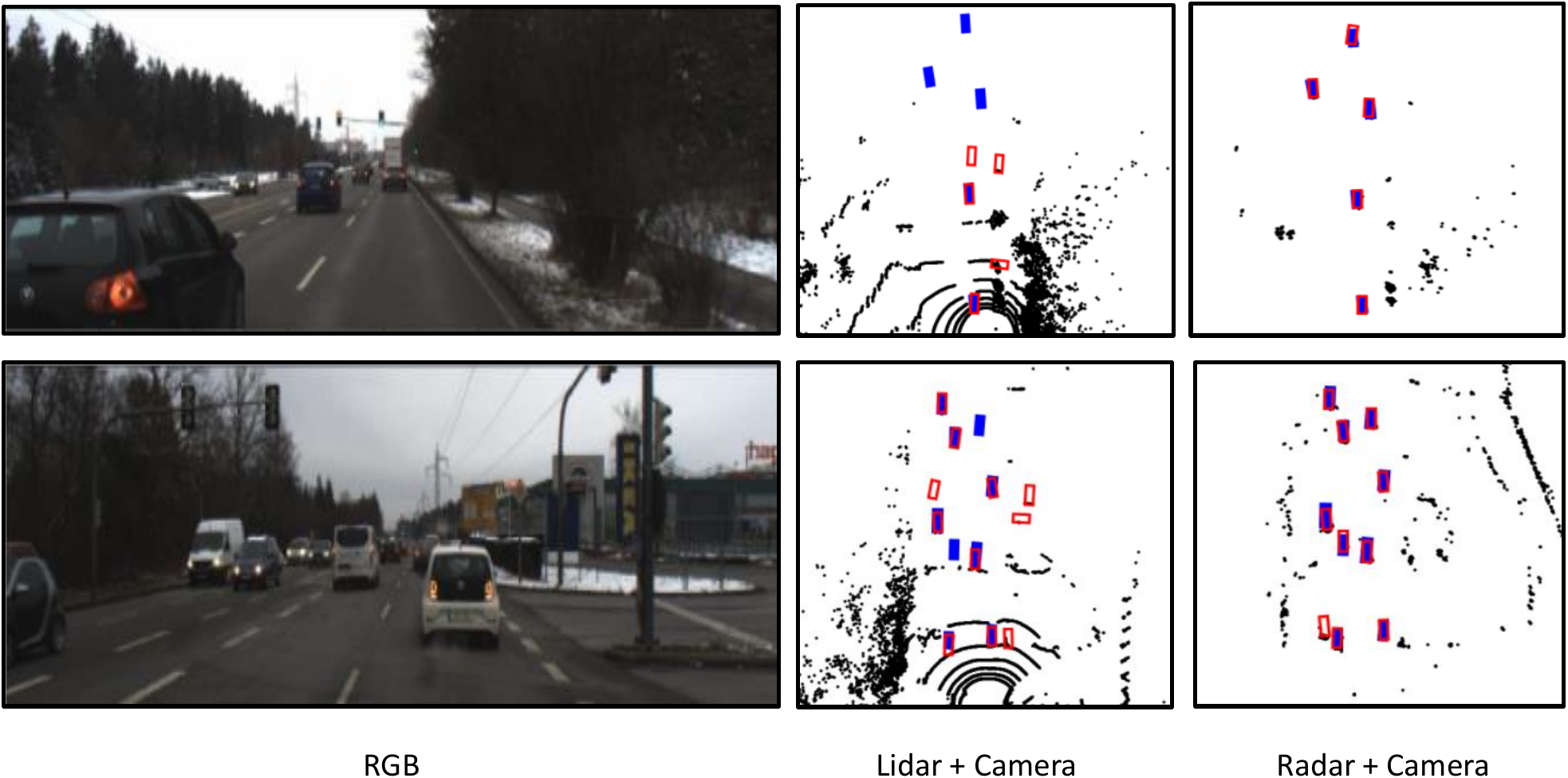}
    \caption{Visualization of the output bounding box output on Astyx Dataset while using lidar and radar data. Black dots represent point clouds in BEV. Filled blue boxes \textcolor{blue}{$\blacksquare$} are ground truth and red empty boxes \textcolor{red}{$\square$} are prediction results.}
    \label{fig:qual_lidar_vs_radar}
\end{figure*}

In section 6.3, we provided results of bounding box prediction when we use the lidar point clouds as input to RadSegNet. Here, we further analyse results on lidar input to better understand the advantage of using radars. We use the 16-channel lidar data provided in the Astyx dataset~\cite{meyer2019automotive} for this experiment. We analyse how the performance gets affected as the distance from the ego vehicle increases. We train the same network (RadSegNet) with lidar and camera data and compare the results with network trained on radar and camera data. Figure~\ref{fig:lidarvsradar} provides the comparison results for this experiment. The result shows that lidar performs better than radar at closer distances, but the performance drops much more significantly as the distance increases. The reason behind this is in alignment to our hypothesis, that lidars provide uniform point clouds which aids the object detection, specially at smaller distances, but at longer ranges, where lidar point density decreases and more occlusions take place, lidar's performance degrades. On the other hand, radars can operate at much longer ranges and provide reliable results throughout even in cases of occlusions. We also provide some qualitative outputs of the experiment in figure~\ref{fig:qual_lidar_vs_radar}. The figure shows how lidar's performance degrades in cases of occlusion and long range. Note that a lidar with more number of channels can provide more density for point clouds, but the problem of occlusion would still be present.



\section{Bad Weather Implementation Details}
\begin{figure*}[t!]
    \centering
    \includegraphics[width=0.95\linewidth]{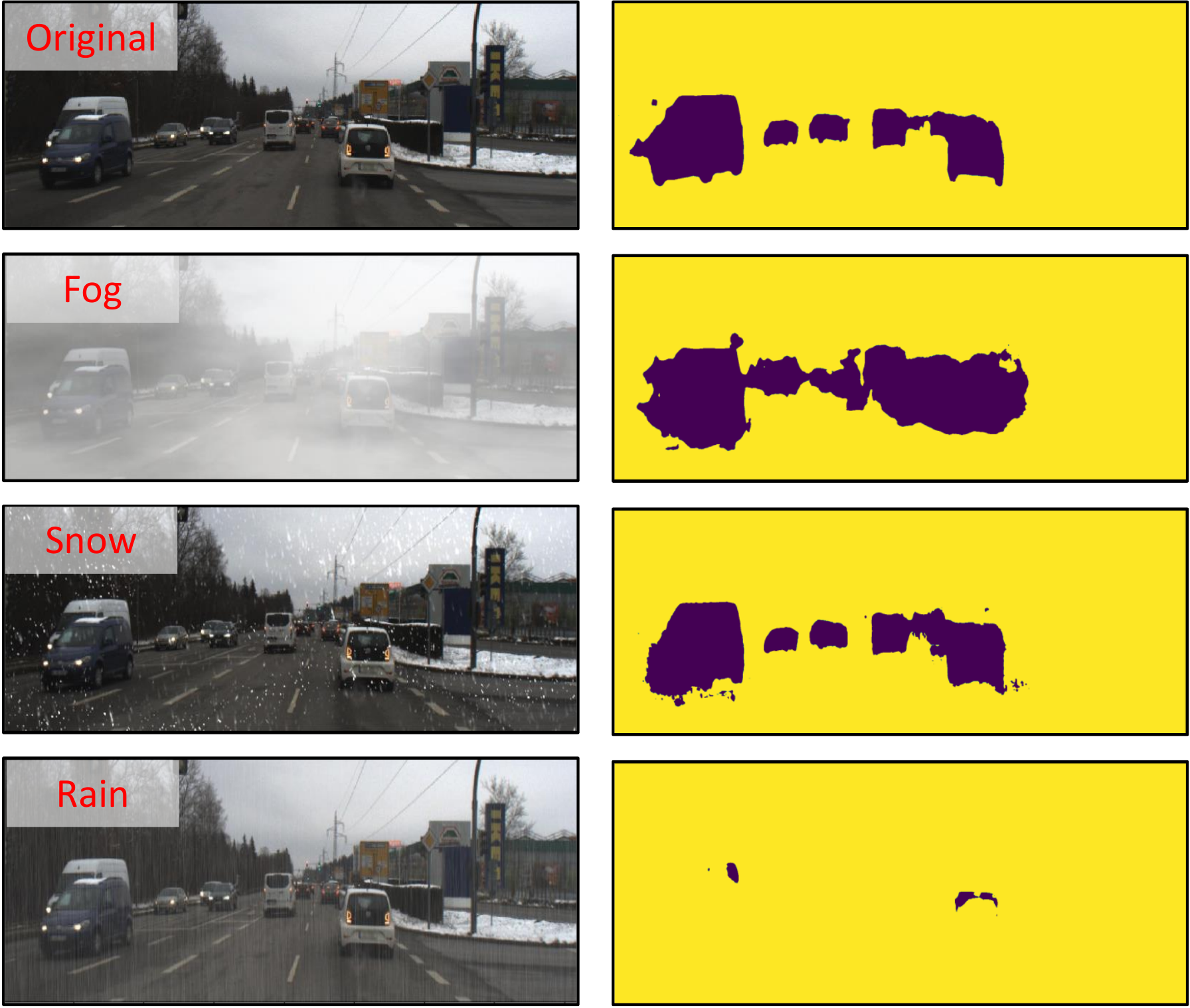}
    \caption{Effect on segmentation outputs for class car with different adversarial weather. Left and right column shows the RGB and corresponding segmentation output.}
    \label{fig:bad_seg}
\end{figure*}
In section 6.3, we provided analysis of the performance of different architectures on bad weather data. We specifically considered most commonly encountered conditions of fog, snow and rain. As the dataset does not provide dense depth maps for physically motivated weather augmentations~\cite{halder2019physics,sakaridis2018semantic}, we used imgaug \footnote{https://imgaug.readthedocs.io/en/latest/source/overview/weather.html} library which simulates bad weather conditions using image filters. For fog, we use the method $fog()$ with argument $seed=5$. For rain, we use $Rain()$ with $drop\_size=(0.10, 0.20)$. For snow we use $SnowFlakes()$ with $flake\_size=(0.7, 0.95), speed=(0.001, 0.03)$. Figure~\ref{fig:bad_seg} shows a sample output of each of these augmentations. We chose these parameters to obtain the perceptually best augmentations.  
We also show the segmentation output of the semantic segmentation network\cite{chen2018encoder} on each of these augmentations. In, snowy conditions segmentation output is not affected heavily, while in foggy conditions we see a lot of false segmentations around the objects. In both these cases RadSegNet retains an almost perfect performance as in clear weather. For rain, segmentation output is most significantly affected. This results into some loss in performance. Nevertheless, even with the affected segmentations, RadSegNet maintains a much more reliable performance compared to state-of-the-art~\cite{kim2020low} camera-radar fusion approach as it learns the features independently from both modalities.

\section{Visualizations on Astyx}
\begin{figure*}[t!]
    \centering
    \includegraphics[width=0.95\linewidth]{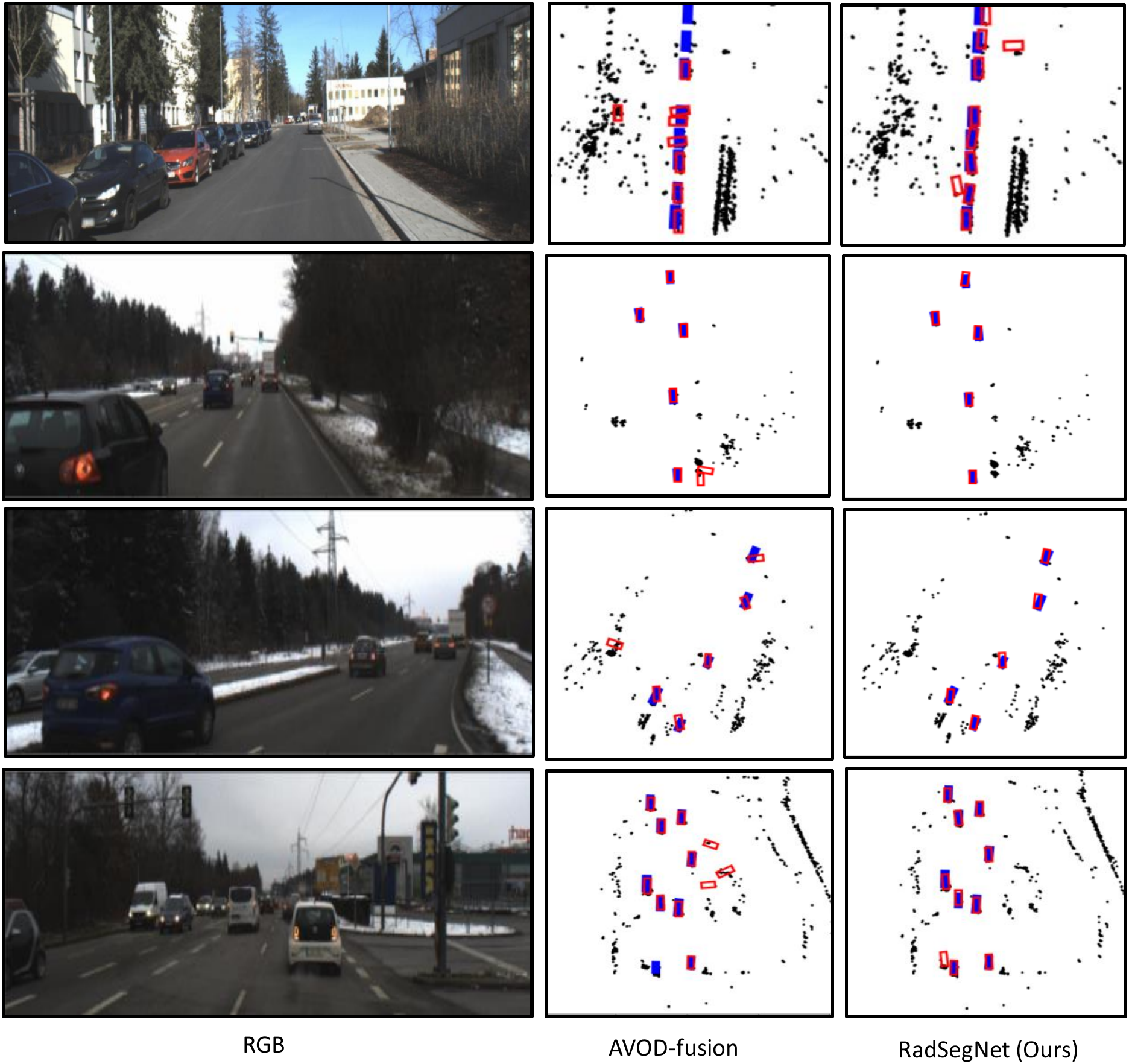}
    \caption{Visualization of the output bounding box output on clear Astyx Dataset. Black dots represent radar point clouds in BEV. Filled blue boxes \textcolor{blue}{$\blacksquare$} are ground truth and red empty boxes \textcolor{red}{$\square$} are prediction results.}
    \label{fig:qual_suppl_astyx}
\end{figure*}

\begin{figure*}[t!]
    \centering
    \includegraphics[width=0.95\linewidth]{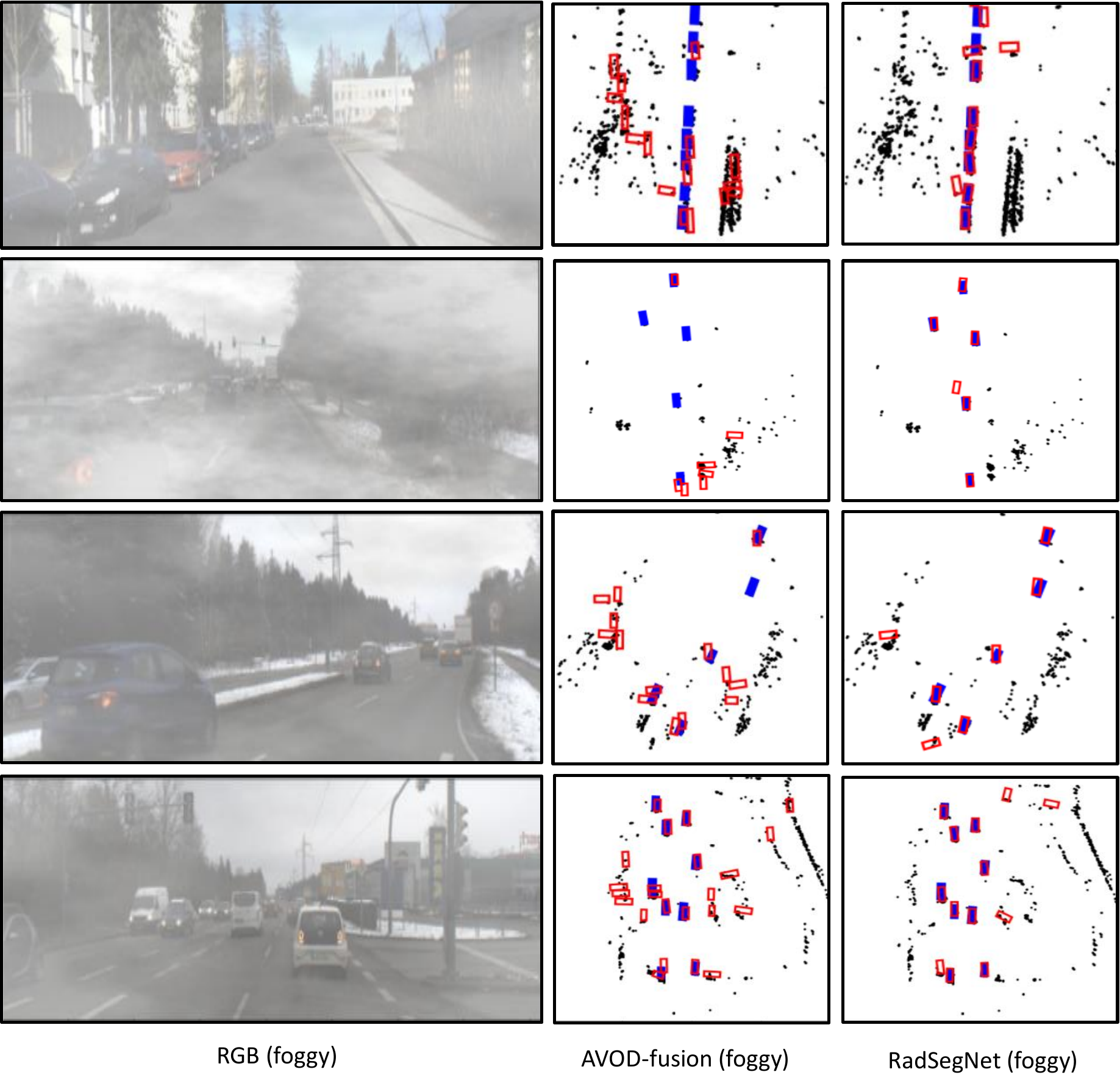}
    \caption{Visualization of the output bounding box output on foggy Astyx Dataset. Black dots represent radar point clouds in BEV. Filled blue boxes \textcolor{blue}{$\blacksquare$} are ground truth and red empty boxes \textcolor{red}{$\square$} are prediction results.}
    \label{fig:qual_suppl_astyx_fog}
\end{figure*}

\begin{figure*}[t!]
    \centering
    \includegraphics[width=0.95\linewidth]{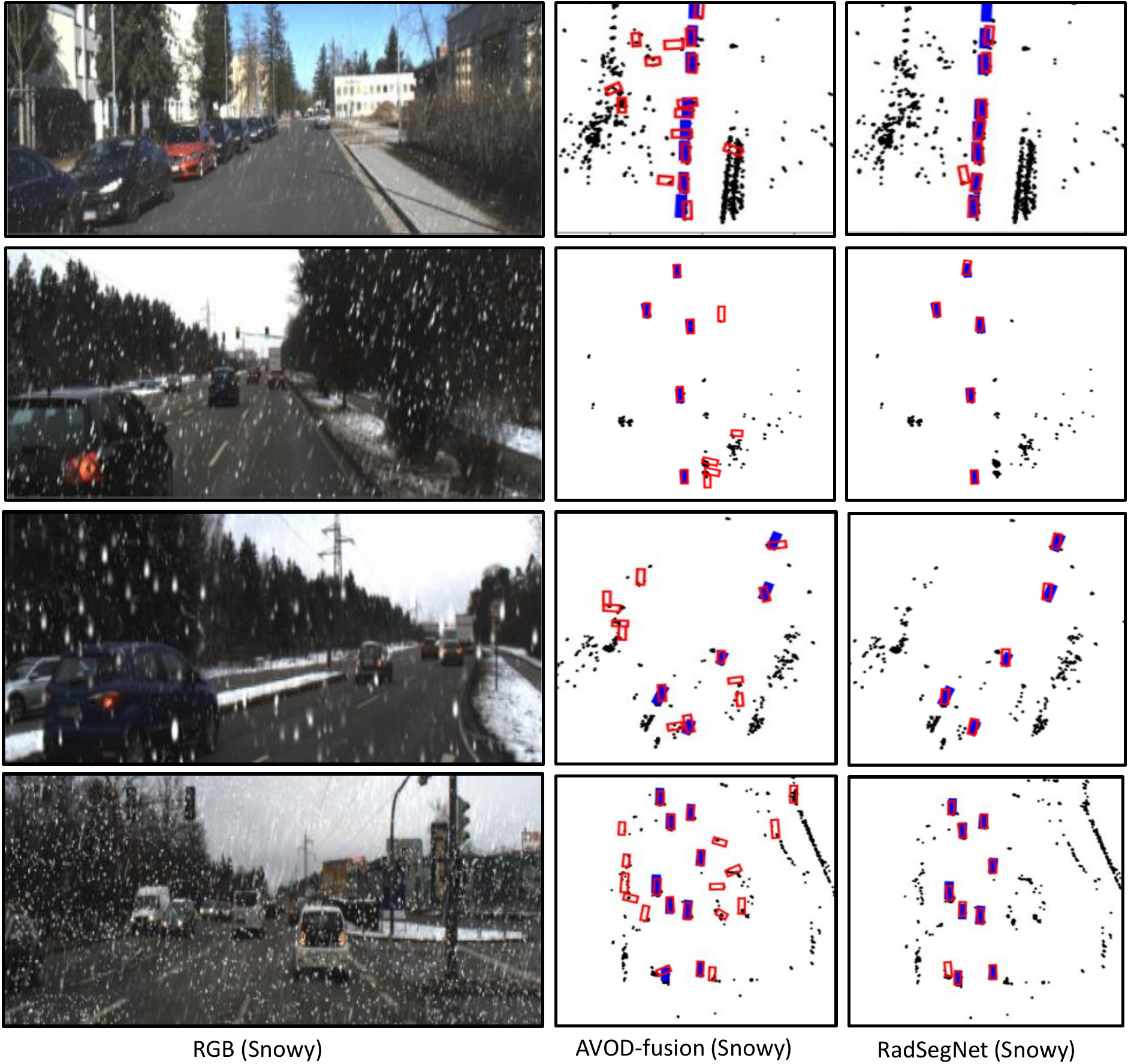}
    \caption{Visualization of the output bounding box output on snowy Astyx Dataset. Black dots represent radar point clouds in BEV. Filled blue boxes \textcolor{blue}{$\blacksquare$} are ground truth and red empty boxes \textcolor{red}{$\square$} are prediction results.}
    \label{fig:qual_suppl_astyx_snow}
\end{figure*}

\begin{figure*}[t!]
    \centering
    \includegraphics[width=0.95\linewidth]{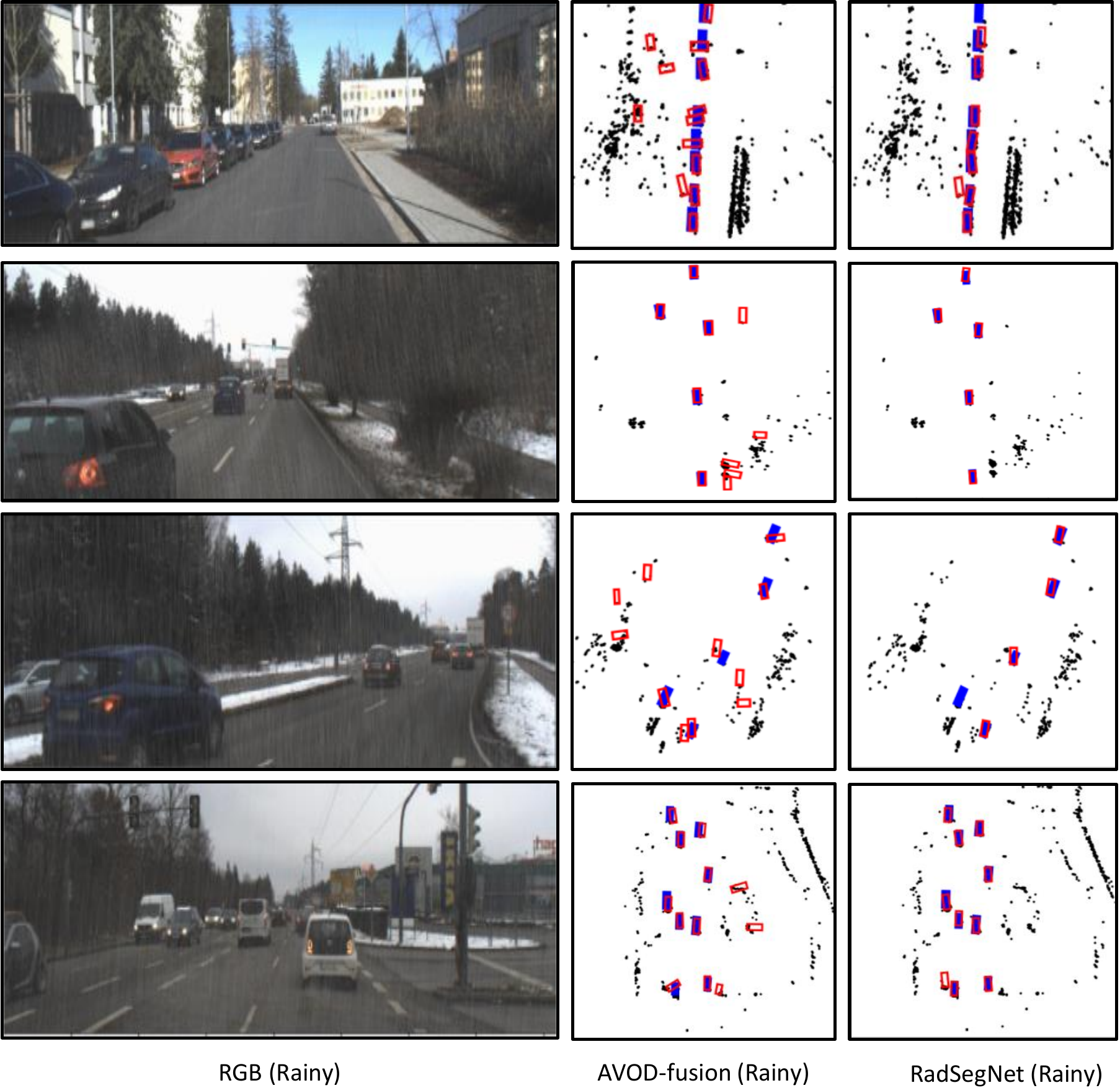}
    \caption{Visualization of the output bounding box output on rainy Astyx Dataset. Black dots represent radar point clouds in BEV. Filled blue boxes \textcolor{blue}{$\blacksquare$} are ground truth and red empty boxes \textcolor{red}{$\square$} are prediction results.}
    \label{fig:qual_suppl_astyx_rain}
\end{figure*}

Figure~\ref{fig:qual_suppl_astyx} provides qualitative bounding box prediction output of RadSegNet compared to AVOD-fusion~\cite{kim2020low}. We choose examples comprising of multiple challenging scenarios from the dataset including long range and occlusions, which shows that the dataset contains enough variability for a comprehensive evaluation. In all the above cases, RadSegNet performs much better than AVOD-fusion, both in terms of accurate detections and less false positives. We also provide visualization of the performance of RadSegNet compared to the baseline (both trained on good weather data only) in different weather conditions (figure~\ref{fig:qual_suppl_astyx_fog},\ref{fig:qual_suppl_astyx_snow},\ref{fig:qual_suppl_astyx_rain}). While the multi-view aggregation based fusion approach gets severely affected in foggy environment, RadSegNet continues to provide reliable detections thanks to its reduced dependence on unreliable camera features. 

\section{Architectural details}
\begin{figure*}[t!]
    \centering
    \includegraphics[width=\linewidth]{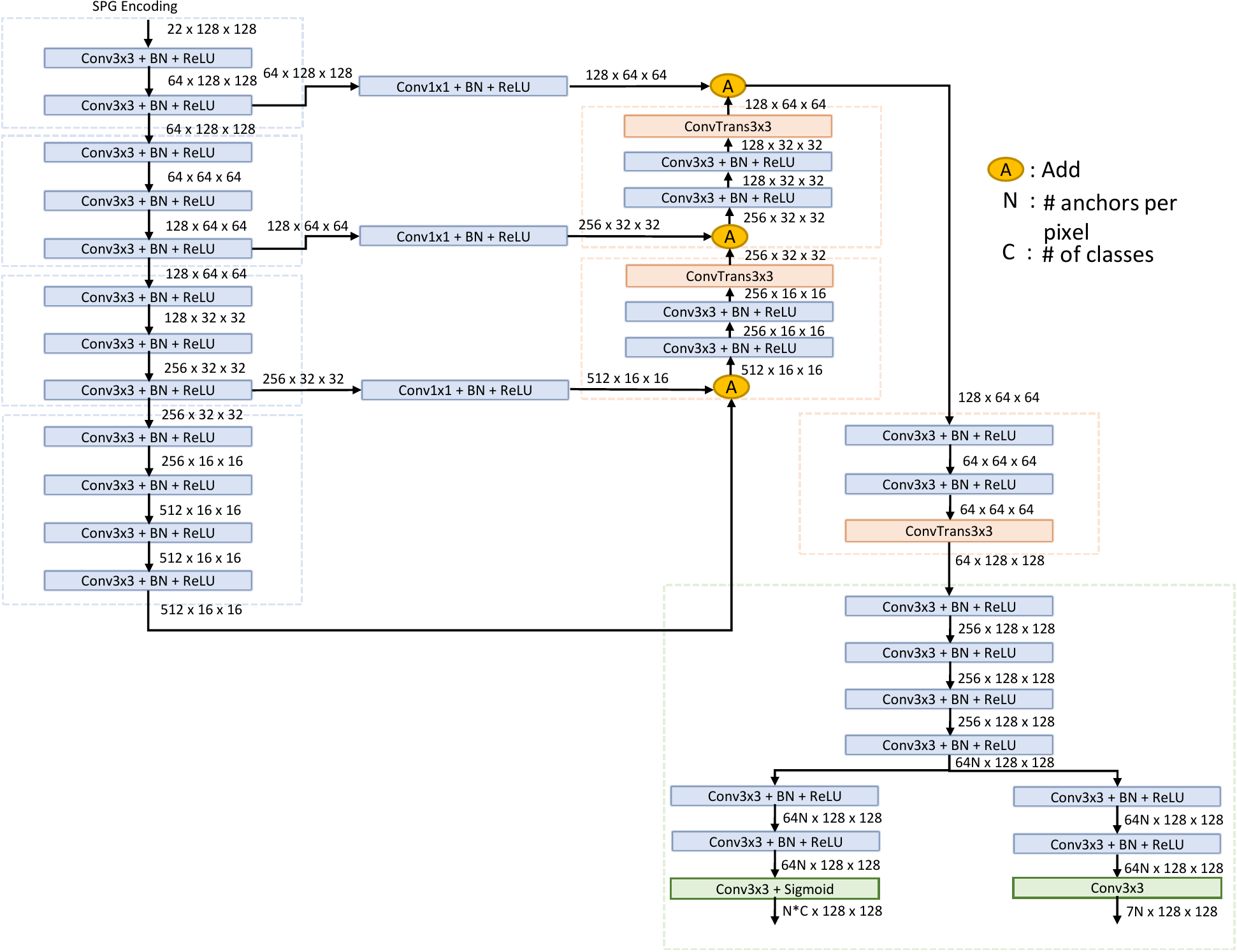}
    \caption{Detailed architecture of RadSegNet}
    \label{fig:arch_details}
\end{figure*}

We present the detailed architecture of RadSegnet in figure~\ref{fig:arch_details}. RadSegnet is a single-stage object detector that adopts a U-net style architecture where features are extracted from SPG encoded input using convolutional layers. There are 3 stages of downsampling in our network and corresponding 3 stages of upsampling. We use stride of 2 for downsampling and transposed convolutions for upsampling. Skip connections between downsampling and upsampling features ensure propagation of finer resolution features. The obtained features are passed to a detection network with two separate heads for classification score prediction and bounding box parameter regressions. 

\end{document}